\documentclass[final]{cvpr}

\usepackage{times}
\usepackage{epsfig}
\usepackage{graphicx}
\usepackage{amsmath}
\usepackage{amssymb}
\usepackage{multirow}
\usepackage{array}

\usepackage[pagebackref=true,breaklinks=true,colorlinks,bookmarks=false]{hyperref}

\def\cvprPaperID{4683} 
\def\confYear{CVPR 2021}

\begin{document}

\title{Multi-Scale Progressive Fusion Learning for Depth Map Super-Resolution}

\author{Chuhua Xian, Kun Qian, Zitian Zhang\\
School of Computer Science and Engineering,\\
South China University of Technology\\
South Campus(HEMC), South China University of Technology, Guangzhou, China\\
{\tt\small chhxian@scut.edu.cn}
\and
Charlie C.L. Wang\\
University of Manchester\\
George Begg Building, Sackville Street, Manchester M13 9PL, UK\\
{\tt\small  changling.wang@manchester.ac.uk}
}

\maketitle

\begin{abstract}
   Limited by the cost and technology, the resolution of depth map collected by depth camera is often lower than that of its associated RGB camera. Although there have been many researches on RGB image super-resolution  (SR), a major problem with depth map super-resolution is that there will be obvious jagged edges and excessive loss of details. To tackle these difficulties, in this work, we propose a multi-scale progressive fusion network for depth map SR, which possess an asymptotic structure to integrate hierarchical features in different domains. Given a low-resolution (LR) depth map and its associated high-resolution (HR) color image, We utilize two different branches to achieve multi-scale feature learning. Next, we propose a step-wise fusion strategy to restore the HR depth map. Finally, a multi-dimensional loss is introduced to constrain clear boundaries and details. Extensive experiments show that our proposed method produces improved results against state-of-the-art methods both qualitatively and quantitatively.
\end{abstract}

\section{Introduction}
\label{introduction}

Depth maps have been widely used in computer graphics and computer vision.
Recently, improved depth sensors have shown great power in both 3D computer vision research and real-life application such as somatosensory interaction, self-driving vehicles, 3D reconstruction, etc. However, different to RGB cameras, depth sensors could only provide depth maps of limited resolution constrained by cost and current hardware technologies, which makes the reconstructed point cloud too sparse to be used in downstream applications. Moreover, compared with high-resolution(HR) depth maps, low-resolution(LR) depth maps are often lack of plenty of high-frequency information, that is, a lot of depth details are lost. To facilitate the use of depth maps, we usually need depth maps with higher resolution. Therefore, tasks such as depth up-sampling are often regarded as recovering HR depth maps from LR depth maps.

Although there have been lots of work on color image SR, depth map SR is still an ill-inverse problem with many challenges.
Previous researchers present optimization-based~\cite{diebel2006application,yang2007spatial,he2010guided,liu2013joint} and  filtering-based methods~\cite{petschnigg2004digital,kopf2007joint,hornacek2013depth,li2016deep} to address this problem. In the framework of optimization methods, depth up-sampling is usually regarded as a process of solving global optimization equations. The optimization formula generally consists of data items, smoothing items and regular items. These methods based on hand-designed objective functions can perform well for recovering HR depth maps in terms of visual quality. But the ability of these methods to capture the global structure is weak, and the computation is usually time-consuming. Another type of filtering-based methods takes more into consideration the design of the spatial domain and the value domain. They utilize the guidance of HR color maps to filter depth images. However, these methods will cause obvious edge aliasing and excessive loss of details, and may introduce texture artifacts when the color image bring biased guidance. In recent years, great success has been made in super-resolution(SR) of RGB images by using very deep convolutional neural networks (CNNs)~\cite{dong2015image,ledig2017photo,zhang2018residual}. Some researchers transfer such CNN models to depth images SR~\cite{hui2016depth,peng2017depth,voynov2019perceptual}.These methods usually fuse the semantic information of input HR images and the features of LR depth maps to generate HR depth maps. Due to the impressive performance of CNNs in image perception, a number of works get well results in depth map SR. Compared with RGB images, depth images express the relative position of objects in three-dimensional (3D) space, so the edges between the surfaces of different and distant objects are clear and sharp. On the contrary, the traditional CNN model often interpolate transitionally at edges, which brings additional noises at edges. Besides, there is a further problem that the most existed networks do not fully integrate the features of the LR depth map and the corresponding guidance image, which causes the depth pixels of the recovery are often blurred. Therefore, it is quite a challenge to reconstruct HR depth maps.

In this paper, to effectively tackle the problems mentioned above, we firstly propose a network framework for depth up-sampling with multi-scale fusion modules. Our network mainly consists of three parts: two encoder parts with skip connections and fusion branch for RGB-D pairs, and a recovery decoding module. Given an input HR color image associated to the LR depth map, a relatively deep block based on current influential backbones is proposed to extract different levels of features. Then we extract features of the LR depth map through constructing pyramid encoder structure. In addition, we merge the features of the first two encoders and the features of the previous stage by a fusion module. Finally, we restore the HR depth map from the learned feature map. Furthermore, we introduce a boundary metric from the traditional image processing field to evaluate the quality of the depth map. This condition improves the generated HR depth map, which makes it possible to obtain boundaries with more details.

The rest of this paper is organized as follows. Section~\ref{relatedwork} review related works on depth SR. Section~\ref{methods} describes the details of our proposed method. In Section 4 and Section 5, we respectively conduct plentiful experiments and analyze the results. Finally, we conclude our work in Section~\ref{conclusions}.

\section{Related Work}
\label{relatedwork}
According to the different starting points and solutions, the related work of depth map SR can be classified into four categories: local depth map SR, global depth map SR, dictionary-based depth map SR methods, and learning-based depth map SR methods.

\subsection{Local Depth Map SR Methods}
Local methods usually consider the use of HR color images guidelines, and local pixel relationships to do up-sampling for LR depth maps. Joint Bilateral Up-sampling(JBU)~\cite{kopf2007joint} considered the Gaussian distance of HR images and LR images in the spatial domain to up-sample the depth map. Liu~\textit{et al.}~\cite{liu2013joint} extended Kopf's work~\cite{kopf2007joint}, considering the geodesic paths of depth pixels based on joint filtering. Besides, they proposed neighborhood hypothesis and distance hypothesis to speed up the filtering calculation, which achieved real-time performance. Yang~\textit{ et al.}~\cite{yang2007spatial} presented a framework including cost volume and sub-pixel refinement to produce a HR depth map. Choi~\cite{choi2014consensus} proposed different up-sampling strategies for continuous and discontinuous regions in the depth map. For depth discontinuous areas, the depth-histogram-based method they proposed made the recovered depth boundary sharper. Lu~\cite{lu2015sparse} proposed to utility the guidance of image segmentation and boundary as a priority for depth up-sampling, which made the boundaries of the HR depth map more clear.

\subsection{Global Depth Map SR Methods}
This type of method usually considers the correlation between color images and depth maps and treat the depth SR task as a global optimization problem on this basic. Diebel~\cite{diebel2006application} was the first to apply Markov Random Fields(MRF) to generate HR depth maps, which considered the constraints of potential distance terms and depth smoothing terms between LR depth map and HR intensity image. Xie~\textit{et al.}~\cite{journals/tip/XieFS16} introduced the self-similarity and the guidance of HR edge map for depth super resolution on the basis of MRF, which also achieved better results. Park~\textit{et al.}~\cite{conf/iccv/ParkKTBK11} proposed an optimization framework for depth SR. It solved the objective function of depth upsampling by taking data items, smoothing items and anisotropic structural-aware items as regular items. Ferstl~\textit{et al.}~\cite{ferstl2013image} proposed anisotropic operators to solve the optimization problem of depth upsampling. Schall~\textit{et al.}~\cite{schall2007feature} introduced the similarity of non-local blocks of HR images and LR images for depth upsampling. Huhle~\textit{et al.}~\cite{huhle2008robust} extended Schall's work~\cite{schall2007feature}, proposing to integrate the local block information of color maps and self-similarity in depth maps to tackle the boundary discontinuities for HR depth generation. Li~\textit{et al.}~\cite{li2016fast} proposed a cascaded global interpolation framework to recover the HR depth map. To some extent, this cascading structure can reduced the texture-copy artifacts and over-smoothing around weak edges caused by color map guidance. The above methods can restore HR depth maps in some situations, but they are time-consuming and cannot fully simulate the correct HR depth distribution in most cases.

\subsection{Dictionary Depth Map SR methods}
This type of method finds the potential relationship between LR and HR image pairs through sparse coding. Yang~\textit{et al.}~\cite{yang2010image} proposed a method to solve the coefficients of a dictionary in LR images to generate HR images. Zheng~\textit{et al.}~\cite{zheng2013depth} introduced a multi-dictionary sparse representation and an adaptive dictionary selection strategy to make the coefficients of HR depth maps more accurate. Kiechle~\textit{et al.}~\cite{kiechle2013joint} treat the depth map SR as a linear inverse problem. And they presented a bimodal co-sparse analysis model to find the interdependency of registered intensity and depth information, which is used to jointly reconstruct HR depth maps. Kwon~\textit{et al.}~\cite{kwon2015data} proposed a data-driven approach to generate HR depth maps through multi-dictionary sparse representation. Their results can solve the problem of over-smoothing on the recovered depth.


\begin{figure*}[!h!t]
  \centering
  \includegraphics[width=0.9\textwidth]{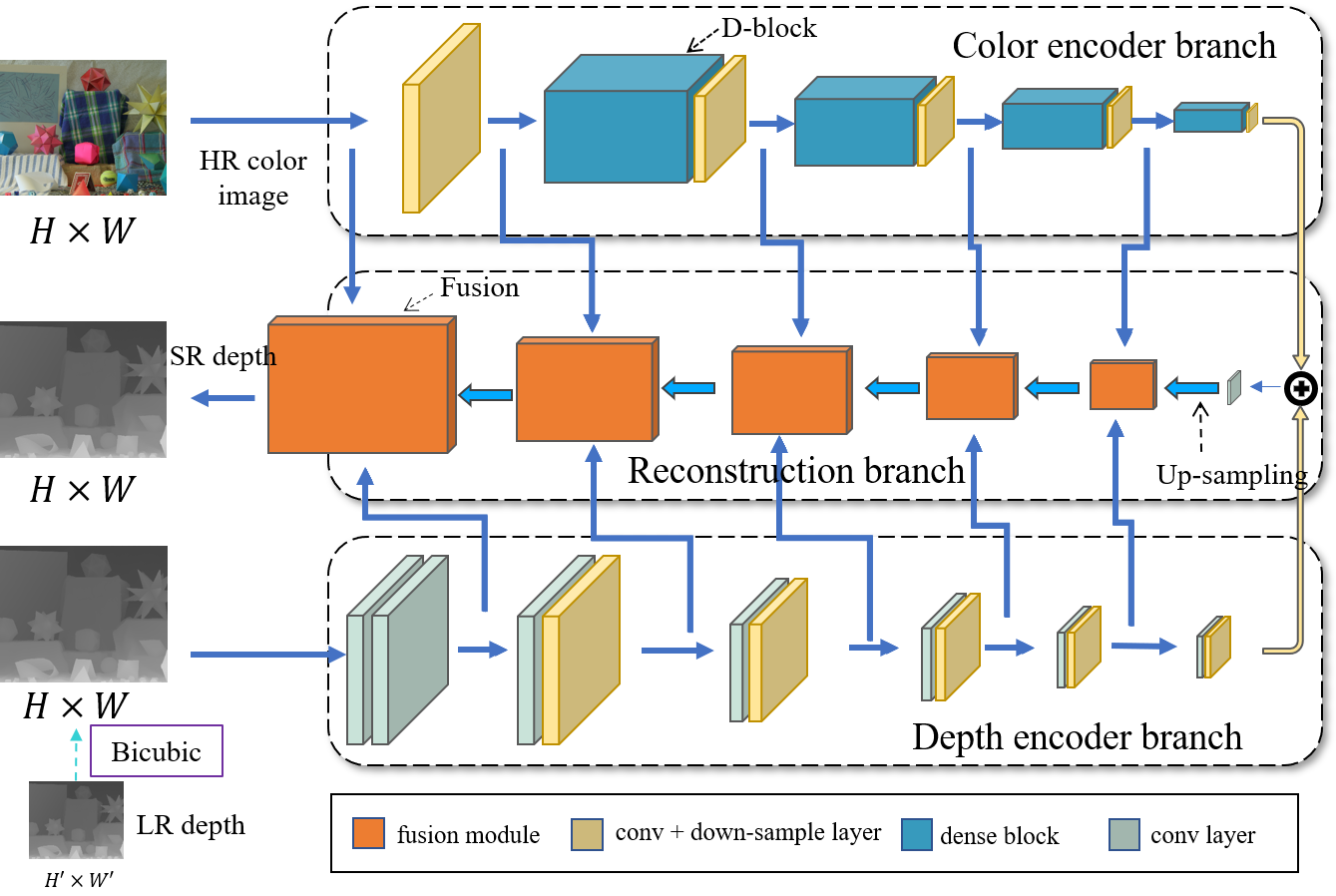}
  \caption{An overview of the proposed framework. The network is based on an encoder-decoder framework with multi-scale feature fusion. From left to right, there are two branches respectively for the progressive feature extraction of the color map and the depth map with multiple level receptive fields. In the reconstruction branch, the features of the color map and the previous hierarchical features continue to guide the restoration of the depth image.}
  \label{framework}
\end{figure*}

\subsection{Learning-based Depth Map SR methods}
With the development of deep learning-based methods in image processing, many learning-based SR methods have also been extensively developed. Dong~\textit{et al.}~\cite{dong2015image} found the underlying relationship between HR images and LR images through deep CNN network, which provided a non-linear mapping learning ability between image pairs. Hui~\textit{et al.}~\cite{hui2016depth} introduced a multi-scale network for depth SR. This network, which used the HR intensity image as guidance, could obtain a HR depth image with up-scaling factors of $\times16$. Riegler~\textit{et al.}~\cite{riegler2016deep} introduced variational optimization on the basis of deep CNN to guide the generation of HR depth maps. Guo~\textit{et al.}~\cite{guo2018hierarchical} presented to learn residuals at different resolutions to guide LR depth maps for accurate interpolations. Voynov~\textit{et al.}~\cite{voynov2019perceptual} proposed perceptual metrics to constrain the network to recover HR depth maps. Experiments had proved that this kind of quality measures which is similar to human perception is more reasonable. Wang~\textit{et al.}~\cite{wang2020depth} proposed a cascaded restoration network, which considered the edge and color information of the input image. Experimental results showed that restoration module including edge information improved the boundaries resolution of recovered depth images.

\section{Proposed Method}
\label{methods}

In this section, we first introduce the framework of our network proposed in this work. Then we briefly present the feature fusion module and how they improve the ability of network to learn mapping relationships. At the end, we will present the loss function and the implementation details.

\subsection{Overview}
Since the degradation forms of HR depth map to LR depth map are infinite in theory, we propose a deep convolutional network to find the optimal SR solution of the non-linear mapping. As shown in Fig.~\ref{framework}, the architecture of our proposed method mainly consists of three branches: the~\textbf{depth encoder branch}, the~\textbf{reconstruction branch} and the~\textbf{color encoder branch}. The two encoder branches with multi-level receptive fields produce a set of hierarchical deep features. Suppose the input resolution of these two encoders is $H \times W$. For the depth encoder branch, Given a LR depth map $D^{{H}' \times {W}'}$, where ${H}'$ and ${W}'$ are obtained according to a certain down-sampling factor $s$ (\textit{e.g.}, $s = 2, 4, 8,$ or $16$). We firstly upscale the LR depth map to the specified resolution using bicubic interpolation. Subsequently, we use this up-scaled depth map as the input of the depth encoder branch. Simultaneously, the associated HR color image $I^{H \times W}$ is input to the color encoder branch which guiding the learning of
rich features in different hierarchies. Notably, compared with the previous encoder, we adopt several dense backbone blocks and residual learning strategy on the color map encoder to fully extract features of the texture structure. Then, we merge and restore the features of different levels in the reconstruction branch. In this branch, a series of fusion modules are set to fuse features of different scales from coarse to fine. In addition, we not only consider the pixel-wise loss between the output and the ground truth, we also take the structural consistency constraint and edge loss into consideration, which will make the recovered HR depth more consistent and the edges become sharper.


\subsection{Network Architecture}
As shown in Fig.~\ref{framework}, our proposed network is based on the encoder-decoder structure as a whole. In the encoding process, it considers different feature extraction strategies for different inputs. During the recovery process, it concatenate hierarchical features through the skip-connections and fusion modules. We will describe each branch in details as follows.

\subsubsection{Color Encoder Branch}
Since the color image itself has the characteristics of rich information and high-resolution color images are easier to obtain, We extract the structural features in the color branch to guide the recovery of HR depth maps. We firstly employ a $3\times3$ convolution and a max-pooling layer with stride $2$ to down-sample the original RGB image, which called \textit{convpool} layer in this work. Moreover, in order to fully extract the hidden features of the color image, we use several dense blocks to further mine the hierarchical features. Following each dense block, we utilize the \textit{convpool} layer which can make the network sensitive to lower-level features. The operations in the color encoder branch can be described as follows:
\begin{equation}
\centering
\begin{aligned}
I^{1,I}_{conv} &=\sigma\left(\mathbf{W}^{1} * I^{1,I}+\mathbf{b}^{1}\right)
\\
F^{1,I}_{down} &= maxpool(I^{1,I}_{conv})
\\
F^{i+1,I}_{down} &= convpool(Dense(F^{i,I}_{down}))
\end{aligned}
\label{equ:colorbranch}
\end{equation}
where $i \in\{1,2,3,4\}$ represents the $i$-th layer, $I^{1,I}_{conv}$ is the $3 \times 3$ convolution result of the input color image $I^{1,I}$ in the color branch. Respectively, $\mathbf{W}^{1,D}$ and $\mathbf{b}^{1,D}$ represent weight and bias in the first convolution in the branch. $\ast $ represents convolution operation and $\sigma$ denotes the element-wise activation function, which adopts rectified linear unit (ReLU). $F^{1,I}_{down}$ is the first output in color branch. $Dense$ denotes dense block extracted based on backbone of DenseNet-121~\cite{huang2017densely} and we apply $convpool$ after each block. The reason of adopting dense block is that it can fully extract the features of input at different scales. Meanwhile, the feature reuse mechanism in it makes the parameters of the network less while further improves the guidance of the reconstruction.


\subsubsection{Depth Encoder Branch}
The depth encoder branch is similar to the color encoder branch. Different to previous branch, we replace the dense block in color encoder branch by a traditional convolutional layer. This is because the depth map has less channel information than the color image, and more complex feature extraction mechanism will induce the network over-fitting. As illustracted in Fig.~\ref{framework}, we first up-sample the input LR depth map by bicubic interpolation to match the resolution $H \times W$ of the color image. Then, we use two transitional convolutional layers with $3 \times 3$ convolution kernel to generate the input feature map. Subsequently,
we used a series of down-sampling modules to extract multi-level features, which can be expressed as:
\begin{equation}
\centering
\begin{aligned}
F^{i+1,D}_{conv} &=\sigma\left(\mathbf{W}^{i,D} * F^{i,D}_{down}+\mathbf{b}^{1,D}\right)
\\
F^{i+1,D}_{down} &= convpool(F^{i,D}_{conv})
\end{aligned}
\label{equ:depthbranch}
\end{equation}
where $i \in\{1,2,3,4,5\}$. In particular, in order to align with the features of the color branch, we added a down-sampling module in the depth branch. For the first feature map $F^{1,D}_{down}$, we extract it from input depth map by two layers with $3 \times 3$ convolution kernel.

\subsubsection{Reconstruction Branch}
We design reconstruction branch to progressively restore HR depth map from the generated hierarchical features from other two branches. For the convenience of expression, we describe the reconstruction branch from right to left. The highest-level feature in this branch contains the abstract semantic information of the recovered depth map. And it is directly concatenated from the feature maps stems from the last layer of color and depth branches. In each step of the subsequent reconstruction, we further integrate the features between different branches through the fusion module. Indeed, the fusion module enables the network to learn the consistency of features in different domains. We formulate our fusion module through fusing the features $F^{m,D}_{down}$, $F^{n,C}_{down}$ and $F^{i,R}_{up}$. The $F^{m,D}_{down}$, $F^{n,C}_{down}$ are $m$-th feature and $n$-th obtained from the color and depth branches respectively while the feature $F^{i,R}_{up}$ denotes the previous step in reconstruction branch. The fusion strategy can be expressed as:
\begin{equation}
\centering
\begin{aligned}
F^{i+1,R}_{f} &= [F^{i,R}_{up}, F^{m,I}_{down}, F^{m,D}_{down}]
\\
F^{i+1,R}_{conv} &=\sigma\left(\mathbf{W}^{i+1,R}_{f} * F^{i+1,R}_{f} +\mathbf{b}^{i+1,R}_{f}\right)
\\
F^{i+1,R}_{up} &= \sigma\left(\mathbf{W}^{i+1,R} * F^{i+1,R}_{conv}+\mathbf{b}^{i+1,R}\right)
\end{aligned}
\end{equation}
where $i \in\{1,2,3,4\}$, $m = k-i-2$ and $n = k-i-1$. $k = 7$ represents the maximum number of modules in the three branches, including the input modules. $F^{i+1,R}_{f}$ concatenates the feature of the corresponding resolution of the color branch, depth branch, and the previous step of the reconstruction branch. And $\mathbf{W}^{i+1,R}_{f}$ and $\mathbf{b}^{i+1,R}_{f}$ are the convolution parameters corresponding to $F^{i+1,R}_{f}$.
Compared with the three fusion modules in the middle, the first $F^{1,R}_{up}$ and last fusion modules $F^{last,R}_{up}$ are slightly different, \textit{i.e.},

\begin{equation}
\centering
\begin{aligned}
F^{1,R}_{f} &= [F^{last,I}_{down}, F^{last,D}_{down}]
\\
F^{1,R}_{up} &= \sigma\left(\mathbf{W}^{1,R} * F^{1,R}_{f}+\mathbf{b}^{1,R}\right)
\\
F^{last,R}_{f} &= [F^{last-1,R}_{up}, I^{1,I}, F^{1,D}_{down}]
\\
F^{last,R}_{conv} &= \sigma\left(\mathbf{W}^{last,R}_{f} * F^{last,R}_{f} +\mathbf{b}^{last,R}_{f}\right)
\\
F^{last,R}_{up} &= \sigma\left(\mathbf{W}^{last,R} * F^{last,R}_{conv} +\mathbf{b}^{last,R}\right)
\end{aligned}
\label{equ:different}
\end{equation}
here $F^{1,R}_{up}$ represents the first fusion module of the reconstruction branch. It directly processes the features which concatenates the output of the last layer of the depth and color branches. Compared to previous fusion modules, the input of last fusion module $F^{last,R}_{f}$ take the original RGB image $I^{1,I}$ as input.




\subsection{Loss Function}
We define three kinds of losses for optimizing the generated SR depth map. We adopt the $L1$ loss to directly constrain each recovery depth to the SR value, an edge loss to improve the depth boundary and make it sharper, and a structure loss to make the recovery depth map more consistent in structure. Finally, these three losses are mixed with a certain weight to restore high-quality depth maps.

\noindent \textbf{$L1$ Loss.} $L1$ loss is a element-wise loss in previous SR method~\cite{li2019feedback,song2020channel}, which can be defined as:

\begin{equation}
L_{l1}(\Theta)=\frac{1}{N} \left\|\mathcal{F}\left(\mathbf{D}^{LR}, \mathbf{I}^{HR} ; \Theta\right)-\mathbf{D}^{HR}\right\|_{1}
\end{equation}
where $L_{l1}(\Theta)$ denotes the $L1$ loss, $\mathcal{F}$ denotes the mapping function of network and $\Theta$ represents the set of trainable parameters in the network. The LR depth map $\mathbf{D}^{LR}$ and HR depth color map $\mathbf{I}^{HR}$ are the input of our network.

\noindent \textbf{Edge Loss.} We further define loss on depth edge to obtain the boundaries with more details. It is an element-wise edge loss based on gradient, and it can be defined as follows:
\begin{equation}
L_{edge}(\Theta)=\frac{1}{N} \left\|sobel(\mathcal{F}\left(\mathbf{D}^{LR}, \mathbf{I}^{HR} ; \Theta\right))-sobel(\mathbf{D}^{HR})\right\|_{1}
\end{equation}
where $L_{edge}(\Theta)$ represents the edge loss, $sobel$ is a boundary operator in the image processing field. It should be noted that this operator requires a hyper-parameter $k$, namely the size of the sliding window. In our experiments, we set $k = 5$.

\noindent \textbf{Structure Loss.} On the contrary, even with the $L1$ loss and edge loss, the restored SR depth value may still be inconsistent in a certain area, which makes the reconstruction effect in the point cloud space worse. So we adopt the structural loss~\cite{wang2004image,zhao2015loss} to favor some local consistency, which can be expressed as:

\begin{equation}
L_{ssim}(\Theta)=SSIM(\mathcal{F}\left(\mathbf{D}^{LR}, \mathbf{I}^{HR} ; \Theta\right), \mathbf{D}^{HR};k)
\end{equation}
where $L_{ssim}(\Theta)$ represents the structure loss. $k$ is the sliding window size used to calculate the SSIM, and we set $k = 11$ in our implementation.

The final loss is a weighted combination of the three losses above, namely $L_{\text {total}}(\Theta)=\lambda_{1} L_{\text {point}}(\Theta)+\lambda_{2} L_{\text {edge}}(\Theta)+\lambda_{3} L_{\text {ssim}}(\Theta)$. In this work, $\lambda_{1} = 0.1$, $\lambda_{1} = 1$ and $\lambda_{3} = 1$ are set to balance the losses for all the experiments.


\section{Experiments}
In this section, we evaluate the performance of our proposed network against several state-of-the-art(SOTA) SR methods on publicly available datasets from a qualitative and quantitative perspective.

In the implementation of our proposed method, we employ ADAM to optimize the parameters of network. We fixed the learning rate to $1e^{-4}$  in the entire training procedure. We perform our training with PyTorch on a PC with an i7-7700 CPU, 16GB RAM, and a GTX 1080Ti GPU.

\subsection{Datasets}
We used three public datasets in this paper: (1) the NYU v2 dataset~\cite{Silberman:ECCV12}, which is captured by both the RGB and Depth cameras from Kinect for indoor scenes; (2) the Middlebury dataset~\cite{scharstein2002taxonomy,scharstein2007learning,scharstein2014high} contains high-quality
depth maps and color maps, and (3) the MPI Sintel depth dataset~\cite{butler2012naturalistic} provides HR color maps and corresponding depth maps.

Significantly, we conduct all experiments on these datasets with different processing methods. The above three datasets are divided into two types of experiments for evaluation. The first is to conduct on NYU, which means training on NYU v2 Raw Dataset containing more than $500,000$ pairs of indoor scenes. And, we compare our proposed method with SOTA methods on NYU v2 labeled Dataset, which consists of $1,449$ RGB-D images. On this dataset we train the proposed network with batch size $8$ for $5$ epochs to convergence. Then, we follow the work in~\cite{hui2016depth} and select $58$ RGB-D images from MPI Sintel depth dataset, and $34$ RGB-D images from Middlebury dataset. We used $82$ images for training
and $10$ images for validation. In addition, we crop the HR depth map to $128 \times 128$ and perform a
sampling on it with stride of $32$ for scaling factors $2, 4, 8$, and $16$ respectively. We augment each patch by a $90^{\circ}$-rotation. Then there are roughly $700,000$ training patches for each scale. To get the LR depth map, we down-sample full-resolution input patch by bicubic interpolation with the given scaling factor ($2, 4, 8$, and $16$). On this dataset(NYU v2), we train the proposed network with batch size $128$ for $5$ epochs.

\subsection{Evaluation}
In order to better effectively evaluate the performance of our proposed method, we conduct experiments clearly between proposed architecture, the bicubic up-sampling, and several state-of-the-art methods: local method (\textit{i.e.},GF~\cite{he2010guided}), global optimization method (\textit{i.e.},TGV~\cite{7676324}), CNN based color map SR methods (\textit{i.e.}, SRCNN~\cite{dong2015image}, RDN~\cite{zhang2018residual}, SRFBN~\cite{li2019feedback}), CNN based depth map SR methods (\textit{i.e.}, MSG~\cite{hui2016depth}, DU-DEAL~\cite{wang2020depth}, DepthSR~\cite{guo2018hierarchical}, and PDDSR~\cite{voynov2019perceptual}). We conduct the quantitative and qualitative analysis further with four scales (\textit{i.e.}, $2, 4, 8$ and $16$) on the datasets processed in the two ways described above. We adopt Root Mean Squared Error (RMSE) and Peak signal-to-noise ratio (PSNR) to evaluate the performance obtained by our method and other state-of-the-art methods. Subsequently, we further compare the running time to show the performance of our method.
Table~\ref{tab1} to ~\ref{tab4} shows the numeric results of the experiments. In Table~\ref{tab1} and ~\ref{tab23}, the best results are shown in bold and the second is underlined. Moreover, we also list the running time of different methods in these tables.

\begin{figure*}
  \centering
  \includegraphics[width=1\textwidth]{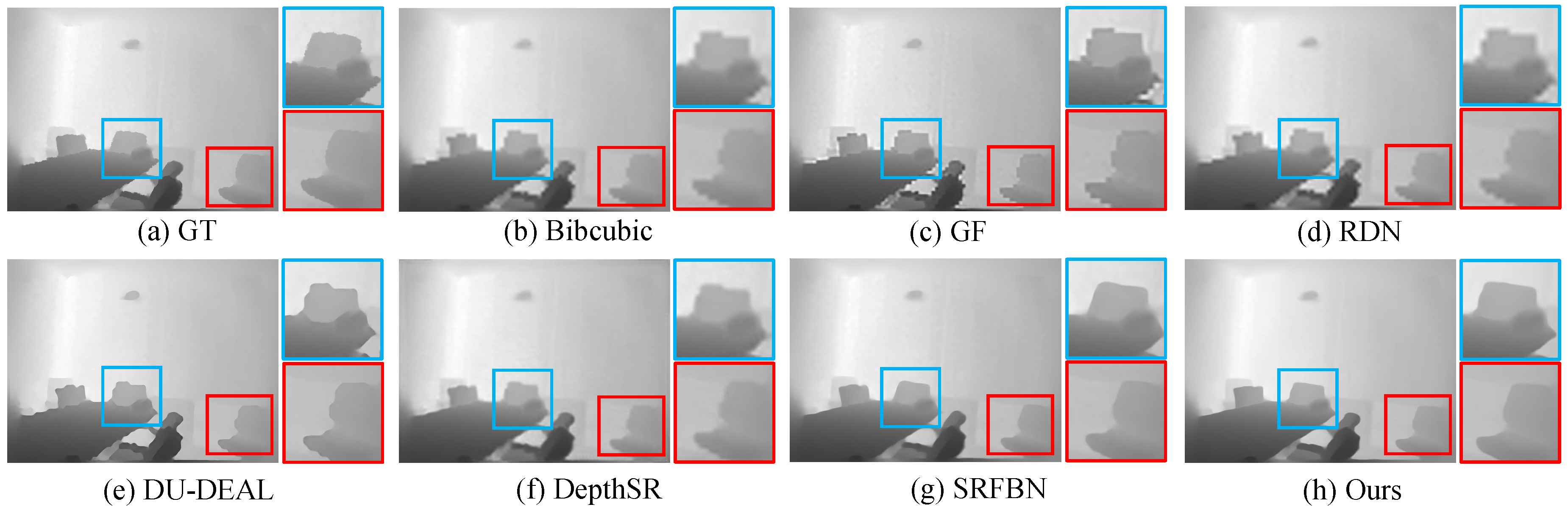}
  \caption{Visual depth SR comparison results for $8\times$ on NYU v2 labeled datasets. From (a) to (h) are the upsampling results of Bibubic, GF~\cite{he2010guided}, RDN~\cite{zhang2018residual}, DU-REAL~\cite{wang2020depth}, DepthSR~\cite{guo2018hierarchical}, SRFBN~\cite{li2019feedback}, and Ours. The details are drawn inside the box.}
  \label{r1}
\end{figure*}

\begin{figure*}
  \centering
  \includegraphics[width=1\textwidth]{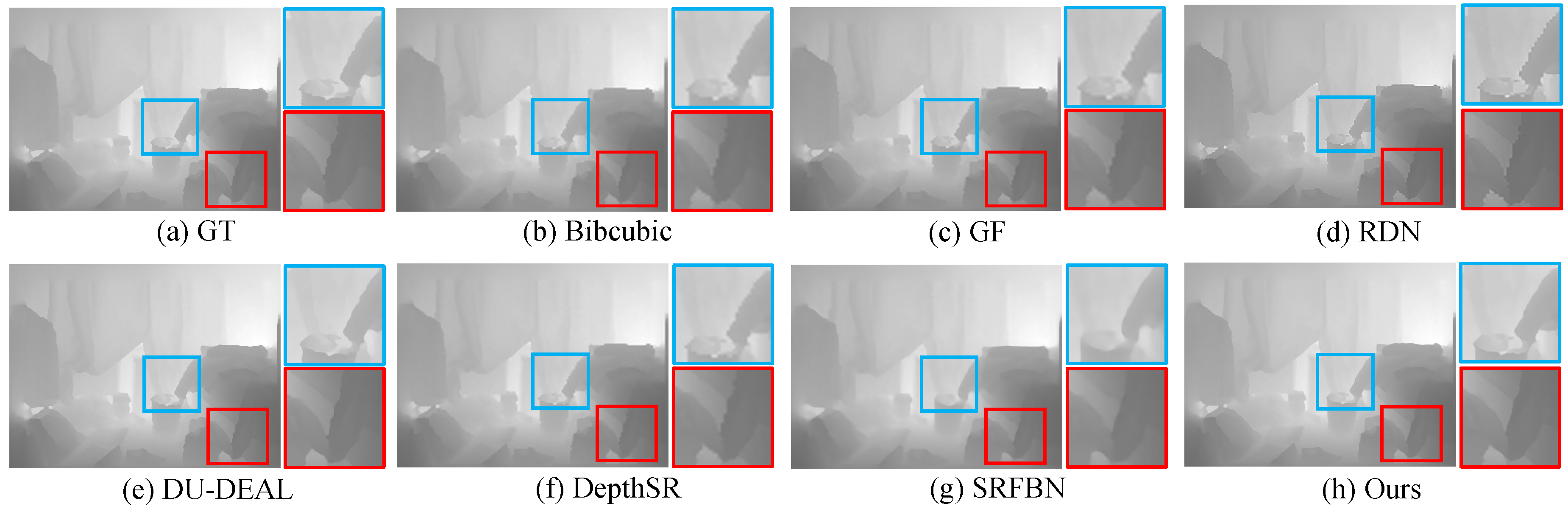}
  \caption{Visual depth SR comparison results for $4\times$ on NYU v2 labeled datasets.}
  \label{r1_2}
\end{figure*}

\noindent \textbf{Evaluation on NYU v2.} We first evaluate the proposed method on NYU v2 dataset. We adopt the publicly available codes of GF~\cite{he2010guided} and TGV~\cite{7676324}. We train RDN~\cite{zhang2018residual}, DepthSR~\cite{guo2018hierarchical}, and SRFBN~\cite{li2019feedback}) on NYU v2 by the authors’s codes and tune to generate best results for evaluation. For MSG~\cite{hui2016depth}, and DU-DEAL~\cite{wang2020depth}), we directly use the released models to depth map SR. We use NYU v2 labeled dataset for testing, which contains $1,449$ RGB-D pairs with $640\times480$ resolution. In order to avoid the impact of missing depth values, we use the Levin's~\cite{Le2004} method firstly to complete all test depth maps. As shown in~\ref{tab1}, the average RMSE/PSNR of our results on this test set is better than all current SOTA methods. Compared with the second best results, our results outperform them of 0.81/1.88 ($2\times$), 0.22/0.35 ($4\times$), 1.7/1.42 ($8\times$), and 2.07/0.77 ($16\times$) of RMSE/PSNR values. Furthermore, we further analyze the performance of our method visually. We perform SR on several depth maps with the down-sampling scale of $4$ and $8$, and the visual results are shown in Fig.~\ref{r1} and Fig.~\ref{r1_2}. As shown in Fig.~\ref{r1_2}, our method not only keeps the boundary details correctly when up-sampling the depth, but also make the restored depth more consistent and reasonable. Additionally, the network we proposed can also recover SR depth at a higher scaling factor. That is, the details of the SR depth map can still be retained under the $\times 8$ scale, as shown in Fig.~\ref{r1}.

In summary, on the real dataset NYU v2,  Bicubic, GF~\cite{he2010guided}, and TGV~\cite{7676324} generate SR results with some artifacts or noises. RDN~\cite{zhang2018residual} and SRFBN~\cite{li2019feedback} are networks designed to restore high-definition color images, which cannot generate the particularly good details when restoring depth maps. DU-DEAL~\cite{wang2020depth} and DepthSR~\cite{guo2018hierarchical} produce competitive results. In contrast, our method can generate the depth boundaries with more details.

\renewcommand\arraystretch{1}
\begin{table*}[ht]
\caption{Quantitative comparisons of four scales on \textbf{NYU v2 labeled Dataset} in terms of average RMSE/PSNR values. The lower the RMSE or the higher the PSNR, the better the performance.}
  \centering
	\begin{tabular}{p{2.8cm}<{\centering}||p{1.2cm}p{1.2cm}p{1.2cm}p{1.2cm}||p{1.2cm}p{1.2cm}p{1.2cm}p{1.2cm}}
		\hline
		\multirow{2}{*}{\textbf{Method}} & \multicolumn{4}{c||}{\textbf{Average RMSE ($\downarrow$)}}                 & \multicolumn{4}{c}{\textbf{Average PSNR ($\uparrow$)}}    \\ \cline{2-9}
		& \textbf{2x}      & \textbf{4x}      & \textbf{8x}      & \textbf{16x}     & \textbf{2x}      & \textbf{4x}      & \textbf{8x}      & \textbf{16x}     \\ \hline
		Bibcubic                          & 4.2              & 4.38             & 6.11             & 7.38             & 39.031           & 36.61            & 33.86            & 31.37            \\ \hline
		GF~\cite{he2010guided}            & 5.41             & 6.07             & 12.64            & 17.18            & 38.03            & 36.23            & 32.31            & 29.25            \\ \hline
		TGV~\cite{7676324}                & 3.2              & 5.18             & 10.11            & 18.09            & 40.05            & 35.91            & 32.17            & 28.17 \\ \hline
		RDN~\cite{zhang2018residual}      & 4.83            & 5.62            & 7.58            & -       & 36.52            & 35.1             & 32.42            & - \\ \hline
		SRFBN~\cite{li2019feedback}       & \underline{2.91}             & \underline{3.79}             & 10.82            & -                & 41.03            & 38.61            & 35.16            & - \\ \hline
		DU-REAL~\cite{wang2020depth}      & 3.08             & 4.47             & 7.19             & 10.32            & \underline{45.47}            & \underline{40.71}            & 35.82            & 31.10            \\ \hline
		DepthSR~\cite{guo2018hierarchical}& 4.23             & 5.2              & \underline{5.53}             & \underline{7.9}             & 40.34            & 37.85            & \underline{37.4}             & \underline{34.05}            \\ \hline
		Ours                              & \textbf{2.1}     & \textbf{3.57}    & \textbf{3.83}    & \textbf{5.83}    & \textbf{47.35}   & \textbf{41.06}   & \textbf{38.82}   & \textbf{34.82}   \\ \hline
	\end{tabular}
	\renewcommand\arraystretch{1.5}
	\label{tab1}
\end{table*}

\begin{figure*}
  \centering
  \includegraphics[width=1\textwidth]{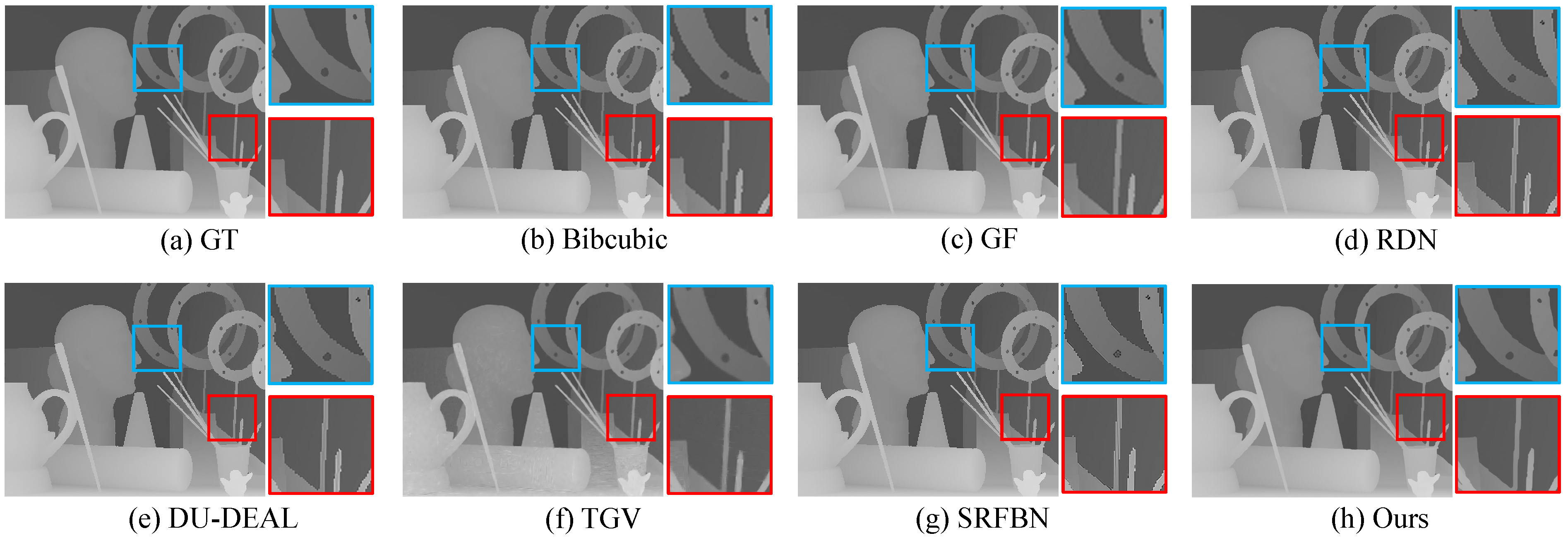}
  \caption{Visual depth SR comparison results for $4\times$ on Middlebury datasets. From (a) to (h) are the HR depth map and the results of Bicubic, GF~\cite{he2010guided}, RDN~\cite{zhang2018residual}, DU-REAL~\cite{wang2020depth}, TGV~\cite{7676324}, SRFBN~\cite{li2019feedback}, and Ours.}
  \label{r2}
\end{figure*}

\textbf{Evaluation on Middlebury.} We carry out experiments on Middlebury dataset. We adopt the publicly available codes of GF~\cite{he2010guided}, TGV~\cite{7676324}, release model of DepthSR~\cite{guo2018hierarchical}, SRFBN~\cite{li2019feedback}), RDN~\cite{zhang2018residual}, DU-DEAL~\cite{wang2020depth}, and we directly use the data in the MSG~\cite{hui2016depth}. For the input of $1320 \times 1080$ resolution, we compare the results using four down-sampling scale factors. The quantitative results are shown in Table~\ref{tab23}. It can be seen from the Table~\ref{tab23} that the RMSE values of our proposed method is based on 6 test data (\textit{i.e.} \textit{Art}, \textit{Books}, \textit{Dolls}, \textit{Laundry}, \textit{Moebius}, \textit{Reindeer}) are generally lower than other methods. We can see that most of the learning-based methods can achieve better depth SR results than traditional methods. In the CNN-based method, the results of methods that focus on depth map SR will generally be better than methods that focus on RGB image SR. And our method will be slightly better than CNN-based depth map SR methods, especially in $\times 4$ and $\times 8$. We also show the visual comparison results in Fig.~\ref{r2}. As shown in Fig.~\ref{r2}, our SR method produces better results in the details (namely in the red box) and boundary (namely in the blue box). Compared with traditional methods, our method also produces less noise. And compared with current learning-based methods, our method can produce more clear and more reasonable SR results of depth maps.

\renewcommand\arraystretch{1.2}
\begin{table*}
\caption{Qualitative analysis results on four scales in terms of RMSE ($\downarrow$) values.}
	\centering
	\begin{tabular}{lllllllllllllllllllllllll}
		\cline{1-13}
		\multicolumn{1}{p{2.5cm}|}{\multirow{2}{*}{\textbf{Method}}} & \multicolumn{4}{c|}{\textbf{Art}}                                                                                                              & \multicolumn{4}{c|}{\textbf{Books}}                                                                                                            & \multicolumn{4}{c}{\textbf{Dolls}}                                                                                       & \multicolumn{4}{c}{\textbf{}}                                                                                                 & \multicolumn{4}{c}{\textbf{}}                                                                                                 & \multicolumn{4}{c}{\textbf{}}                                                                                                 \\ \cline{2-13}
		\multicolumn{1}{c|}{}                                 & \multicolumn{1}{l|}{\textbf{2x}} & \multicolumn{1}{l|}{\textbf{4x}} & \multicolumn{1}{l|}{\textbf{8x}} & \multicolumn{1}{l|}{\textbf{16x}}     & \multicolumn{1}{l|}{\textbf{2x}} & \multicolumn{1}{l|}{\textbf{4x}} & \multicolumn{1}{l|}{\textbf{8x}} & \multicolumn{1}{l|}{\textbf{16x}}     & \multicolumn{1}{l|}{\textbf{2x}} & \multicolumn{1}{l|}{\textbf{4x}} & \multicolumn{1}{l|}{\textbf{8x}} & \textbf{16x}     & \multicolumn{1}{c}{\textbf{}} & \multicolumn{1}{c}{\textbf{}} & \multicolumn{1}{c}{\textbf{}} & \multicolumn{1}{c}{\textbf{}} & \multicolumn{1}{c}{\textbf{}} & \multicolumn{1}{c}{\textbf{}} & \multicolumn{1}{c}{\textbf{}} & \multicolumn{1}{c}{\textbf{}} & \multicolumn{1}{c}{\textbf{}} & \multicolumn{1}{c}{\textbf{}} & \multicolumn{1}{c}{\textbf{}} & \multicolumn{1}{c}{\textbf{}} \\ \cline{1-13}
		\multicolumn{1}{l|}{Bibcubic}                         & \multicolumn{1}{l|}{3.53}        & \multicolumn{1}{l|}{3.84}        & \multicolumn{1}{l|}{4.47}        & \multicolumn{1}{l|}{5.72}             & \multicolumn{1}{l|}{1.31}        & \multicolumn{1}{l|}{1.61}        & \multicolumn{1}{l|}{2.34}        & \multicolumn{1}{l|}{3.34}             & \multicolumn{1}{l|}{3.28}        & \multicolumn{1}{l|}{3.34}        & \multicolumn{1}{l|}{3.47}        & 3.72             &                               &                               &                               &                               &                               &                               &                               &                               &                               &                               &                               &                               \\ \cline{1-13}
		\multicolumn{1}{l|}{GF}                               & \multicolumn{1}{l|}{2.75}        & \multicolumn{1}{l|}{3.91}        & \multicolumn{1}{l|}{5.32}        & \multicolumn{1}{l|}{8.36}             & \multicolumn{1}{l|}{1.36}        & \multicolumn{1}{l|}{1.76}        & \multicolumn{1}{l|}{2.1}         & \multicolumn{1}{l|}{3.36}             & \multicolumn{1}{l|}{1.23}        & \multicolumn{1}{l|}{2.48}        & \multicolumn{1}{l|}{3.97}        & 4.86             &                               &                               &                               &                               &                               &                               &                               &                               &                               &                               &                               &                               \\ \cline{1-13}
		\multicolumn{1}{l|}{TGV}                              & \multicolumn{1}{l|}{3.03}        & \multicolumn{1}{l|}{3.78}        & \multicolumn{1}{l|}{7.08}        & \multicolumn{1}{l|}{11.59}            & \multicolumn{1}{l|}{1.29}        & \multicolumn{1}{l|}{1.61}        & \multicolumn{1}{l|}{2.15}        & \multicolumn{1}{l|}{3.05}             & \multicolumn{1}{l|}{1.63}        & \multicolumn{1}{l|}{1.96}        & \multicolumn{1}{l|}{2.62}        & 4.08             &                               &                               &                               &                               &                               &                               &                               &                               &                               &                               &                               &                               \\ \cline{1-13}
		\multicolumn{1}{l|}{RDN}                              & \multicolumn{1}{l|}{2.61}        & \multicolumn{1}{l|}{3.82}        & \multicolumn{1}{l|}{5.87}        & \multicolumn{1}{l|}{-} & \multicolumn{1}{l|}{1.46}        & \multicolumn{1}{l|}{2.01}        & \multicolumn{1}{l|}{3.08}        & \multicolumn{1}{l|}{-} & \multicolumn{1}{l|}{1.25}        & \multicolumn{1}{l|}{1.7}         & \multicolumn{1}{l|}{2.22}        & - &                               &                               &                               &                               &                               &                               &                               &                               &                               &                               &                               &                               \\ \cline{1-13}
		\multicolumn{1}{l|}{SRFBN}                            & \multicolumn{1}{l|}{1.99}        & \multicolumn{1}{l|}{3.02}        & \multicolumn{1}{l|}{3.58}        & \multicolumn{1}{l|}{-} & \multicolumn{1}{l|}{0.54}        & \multicolumn{1}{l|}{1.22}        & \multicolumn{1}{l|}{1.51}        & \multicolumn{1}{l|}{-} & \multicolumn{1}{l|}{1.04}        & \multicolumn{1}{l|}{1.81}        & \multicolumn{1}{l|}{2.06}        & - &                               &                               &                               &                               &                               &                               &                               &                               &                               &                               &                               &                               \\ \cline{1-13}
		\multicolumn{1}{l|}{MSG}                              & \multicolumn{1}{l|}{0.66}        & \multicolumn{1}{l|}{1.47}        & \multicolumn{1}{l|}{2.45}        & \multicolumn{1}{l|}{4.57}             & \multicolumn{1}{l|}{0.37}        & \multicolumn{1}{l|}{0.67}        & \multicolumn{1}{l|}{1.03}        & \multicolumn{1}{l|}{1.6}              & \multicolumn{1}{l|}{0.34}        & \multicolumn{1}{l|}{0.69}        & \multicolumn{1}{l|}{1.05}        & 1.6              &                               &                               &                               &                               &                               &                               &                               &                               &                               &                               &                               &                               \\ \cline{1-13}
		\multicolumn{1}{l|}{DU-REAL}                          & \multicolumn{1}{l|}{0.62}        & \multicolumn{1}{l|}{\underline{1.15}}        & \multicolumn{1}{l|}{\underline{2.15}}        & \multicolumn{1}{l|}{4.32}             & \multicolumn{1}{l|}{0.34}        & \multicolumn{1}{l|}{\underline{0.57}}        & \multicolumn{1}{l|}{1.01}        & \multicolumn{1}{l|}{1.54}             & \multicolumn{1}{l|}{\underline{0.31}}        & \multicolumn{1}{l|}{0.65}        & \multicolumn{1}{l|}{0.98}        & \underline{1.42}             &                               &                               &                               &                               &                               &                               &                               &                               &                               &                               &                               &                               \\ \cline{1-13}
		\multicolumn{1}{l|}{DepthSR}                          & \multicolumn{1}{l|}{\underline{0.53}}        & \multicolumn{1}{l|}{1.2}         & \multicolumn{1}{l|}{2.22}        & \multicolumn{1}{l|}{\underline{3.91}}             & \multicolumn{1}{l|}{\underline{0.31}}        & \multicolumn{1}{l|}{0.6}         & \multicolumn{1}{l|}{\underline{0.89}}        & \multicolumn{1}{l|}{\underline{1.51}}             & \multicolumn{1}{l|}{0.32}        & \multicolumn{1}{l|}{\underline{0.62}}        & \multicolumn{1}{l|}{\underline{0.85}}        & 1.48             &                               &                               &                               &                               &                               &                               &                               &                               &                               &                               &                               &                               \\ \cline{1-13}
		\multicolumn{1}{l|}{Ours}                             & \multicolumn{1}{l|}{\textbf{0.51}}        & \multicolumn{1}{l|}{\textbf{1.05}}        & \multicolumn{1}{l|}{\textbf{2.08}}        & \multicolumn{1}{l|}{\textbf{3.87}}             & \multicolumn{1}{l|}{\textbf{0.3}}         & \multicolumn{1}{l|}{\textbf{0.55}}        & \multicolumn{1}{l|}{\textbf{0.83}}        & \multicolumn{1}{l|}{\textbf{1.47}}             & \multicolumn{1}{l|}{\textbf{0.29}}         & \multicolumn{1}{l|}{\textbf{0.58}}        & \multicolumn{1}{l|}{\textbf{0.8}}         & \textbf{1.35}             &                               &                               &                               &                               &                               &                               &                               &                               &                               &                               &                               &                               \\ \cline{1-13}
		&                                  &                                  &                                  &                                       &                                  &                                  &                                  &                                       &                                  &                                  &                                  &                  &                               &                               &                               &                               &                               &                               &                               &                               &                               &                               &                               &                               \\ \cline{1-13}
		\multicolumn{1}{c|}{\multirow{2}{*}{\textbf{Method}}} & \multicolumn{4}{c|}{\textbf{Laundry}}                                                                                                          & \multicolumn{4}{c|}{\textbf{Moebius}}                                                                                                          & \multicolumn{4}{c}{\textbf{Reindeer}}                                                                                    &                               &                               &                               &                               &                               &                               &                               &                               &                               &                               &                               &                               \\ \cline{2-13}
		\multicolumn{1}{c|}{}                                 & \multicolumn{1}{p{0.85cm}|}{\textbf{2x}} & \multicolumn{1}{p{0.85cm}|}{\textbf{4x}} & \multicolumn{1}{p{0.85cm}|}{\textbf{8x}} & \multicolumn{1}{p{0.85cm}|}{\textbf{16x}}     & \multicolumn{1}{p{0.85cm}|}{\textbf{2x}} & \multicolumn{1}{p{0.85cm}|}{\textbf{4x}} & \multicolumn{1}{p{0.85cm}|}{\textbf{8x}} & \multicolumn{1}{p{0.85cm}|}{\textbf{16x}}     & \multicolumn{1}{p{0.85cm}|}{\textbf{2x}} & \multicolumn{1}{p{0.85cm}|}{\textbf{4x}} & \multicolumn{1}{p{0.85cm}|}{\textbf{8x}} & \textbf{16x}     &                               &                               &                               &                               &                               &                               &                               &                               &                               &                               &                               &                               \\ \cline{1-13}
		\multicolumn{1}{l|}{Bibcubic}                         & \multicolumn{1}{l|}{3.35}        & \multicolumn{1}{l|}{3.49}        & \multicolumn{1}{l|}{3.77}        & \multicolumn{1}{l|}{4.35}             & \multicolumn{1}{l|}{3.28}        & \multicolumn{1}{l|}{3.36}        & \multicolumn{1}{l|}{3.5}         & \multicolumn{1}{l|}{3.81}             & \multicolumn{1}{l|}{3.4}         & \multicolumn{1}{l|}{3.52}        & \multicolumn{1}{l|}{3.83}        & 5.82             &                               &                               &                               &                               &                               &                               &                               &                               &                               &                               &                               &                               \\ \cline{1-13}
		\multicolumn{1}{l|}{GF}                               & \multicolumn{1}{l|}{2.26}        & \multicolumn{1}{l|}{2.67}        & \multicolumn{1}{l|}{3.84}        & \multicolumn{1}{l|}{5.23}             & \multicolumn{1}{l|}{1.92}        & \multicolumn{1}{l|}{2.29}        & \multicolumn{1}{l|}{3.85}        & \multicolumn{1}{l|}{5.22}             & \multicolumn{1}{l|}{2.27}        & \multicolumn{1}{l|}{2.89}        & \multicolumn{1}{l|}{3.98}        & 5.85             &                               &                               &                               &                               &                               &                               &                               &                               &                               &                               &                               &                               \\ \cline{1-13}
		\multicolumn{1}{l|}{TGV}                              & \multicolumn{1}{l|}{2.15}        & \multicolumn{1}{l|}{2.51}        & \multicolumn{1}{l|}{3.82}        & \multicolumn{1}{l|}{6.41}             & \multicolumn{1}{l|}{1.21}        & \multicolumn{1}{l|}{1.65}        & \multicolumn{1}{l|}{2.13}        & \multicolumn{1}{l|}{2.73}             & \multicolumn{1}{l|}{2.41}        & \multicolumn{1}{l|}{2.71}        & \multicolumn{1}{l|}{3.79}        & 7.27             &                               &                               &                               &                               &                               &                               &                               &                               &                               &                               &                               &                               \\ \cline{1-13}
		\multicolumn{1}{l|}{RDN}                              & \multicolumn{1}{l|}{2.53}        & \multicolumn{1}{l|}{3.22}        & \multicolumn{1}{l|}{4.65}        & \multicolumn{1}{l|}{-} & \multicolumn{1}{l|}{1.22}        & \multicolumn{1}{l|}{1.61}        & \multicolumn{1}{l|}{2.39}        & \multicolumn{1}{l|}{-} & \multicolumn{1}{l|}{3.32}        & \multicolumn{1}{l|}{2.93}        & \multicolumn{1}{l|}{4.41}        & - &                               &                               &                               &                               &                               &                               &                               &                               &                               &                               &                               &                               \\ \cline{1-13}
		\multicolumn{1}{l|}{SRFBN}                            & \multicolumn{1}{l|}{1.67}        & \multicolumn{1}{l|}{2.13}        & \multicolumn{1}{l|}{2.28}        & \multicolumn{1}{l|}{-} & \multicolumn{1}{l|}{1.12}        & \multicolumn{1}{l|}{1.43}        & \multicolumn{1}{l|}{1.52}        & \multicolumn{1}{l|}{-} & \multicolumn{1}{l|}{1.63}        & \multicolumn{1}{l|}{2.07}        & \multicolumn{1}{l|}{2.15}        & - &                               &                               &                               &                               &                               &                               &                               &                               &                               &                               &                               &                               \\ \cline{1-13}
		\multicolumn{1}{l|}{MSG}                              & \multicolumn{1}{l|}{0.37}        & \multicolumn{1}{l|}{0.79}        & \multicolumn{1}{l|}{1.51}        & \multicolumn{1}{l|}{2.62}             & \multicolumn{1}{l|}{0.35}        & \multicolumn{1}{l|}{0.66}        & \multicolumn{1}{l|}{1.02}        & \multicolumn{1}{l|}{1.63}             & \multicolumn{1}{l|}{0.42}        & \multicolumn{1}{l|}{0.98}        & \multicolumn{1}{l|}{1.76}        & 2.91             &                               &                               &                               &                               &                               &                               &                               &                               &                               &                               &                               &                               \\ \cline{1-13}
		\multicolumn{1}{l|}{DU-REAL}                          & \multicolumn{1}{l|}{0.35}        & \multicolumn{1}{l|}{\underline{0.76}}        & \multicolumn{1}{l|}{1.49}        & \multicolumn{1}{l|}{2.56}             & \multicolumn{1}{l|}{0.34}        & \multicolumn{1}{l|}{0.62}        & \multicolumn{1}{l|}{0.97}        & \multicolumn{1}{l|}{1.54}             & \multicolumn{1}{l|}{0.39}        & \multicolumn{1}{l|}{\underline{0.95}}        & \multicolumn{1}{l|}{1.61}        & 2.53             &                               &                               &                               &                               &                               &                               &                               &                               &                               &                               &                               &                               \\ \cline{1-13}
		\multicolumn{1}{l|}{DepthSR}                          & \multicolumn{1}{l|}{\underline{0.34}}        & \multicolumn{1}{l|}{0.78}        & \multicolumn{1}{l|}{\underline{1.32}}        & \multicolumn{1}{l|}{\underline{2.26}}             & \multicolumn{1}{l|}{\underline{0.32}}        & \multicolumn{1}{l|}{\underline{0.59}}        & \multicolumn{1}{l|}{\underline{0.92}}        & \multicolumn{1}{l|}{\underline{1.51}}             & \multicolumn{1}{l|}{\underline{0.39}}        & \multicolumn{1}{l|}{0.96}        & \multicolumn{1}{l|}{\underline{1.57}}        & \underline{2.47}             &                               &                               &                               &                               &                               &                               &                               &                               &                               &                               &                               &                               \\ \cline{1-13}
		\multicolumn{1}{l|}{Ours}                             & \multicolumn{1}{l|}{\textbf{0.32}}        & \multicolumn{1}{l|}{\textbf{0.71}}        & \multicolumn{1}{l|}{\textbf{1.21}}        & \multicolumn{1}{l|}{\textbf{2.15}}             & \multicolumn{1}{l|}{\textbf{0.29}}        & \multicolumn{1}{l|}{\textbf{0.56}}        & \multicolumn{1}{l|}{\textbf{0.85}}        & \multicolumn{1}{l|}{\textbf{1.42}}             & \multicolumn{1}{l|}{\textbf{0.35}}        & \multicolumn{1}{l|}{\textbf{0.82}}        & \multicolumn{1}{l|}{\textbf{1.45}}        & \textbf{2.21}             &                               &                               &                               &                               &                               &                               &                               &                               &                               &                               &                               &                               \\ \cline{1-13}
	\end{tabular}
\label{tab23}
\end{table*}

\textbf{Running Time.} We compare the computation time of our method and other methods on different scales. We test on NYU v2 labeled datasets with $640\times480$ resolution and the test environment is set in a PC with an Intel(R) i5-4590 CPU, 16GB RAM, and an NVIDIA GeForce GTX 1060 $6$GB GPU. The average running time for whole datasets is listed in Table~\ref{tab4}. The implementation codes of Bicubic, GF~\cite{he2010guided}, RDN~\cite{zhang2018residual}, SRFBN~\cite{li2019feedback}, and DepthSR~\cite{guo2018hierarchical} are written in Python (implemented using TensorFlow or Pytorch) with GPU acceleration. TGV~\cite{7676324} and DU-REAL~\cite{wang2020depth} is implemented with Matlab with MatConvNet and MSG~\cite{hui2016depth} is implemented with Caffe. The code of our method is implemented with Pytorch and we use GPU to accelerate it. Observing from Table~\ref{tab4}, we can see that the learning-based method is faster than traditional method due to GPU acceleration. Moreover, the running time of our method is faster than that of other methods except bicubic under different scales.

\begin{table}[]
\caption{The average running time with different scales on the NYU v2 labeled datasets.}
\begin{tabular}{l||l|l|l|l}
\hline
Method   & \textbf{2x}    & \textbf{4x}    & \textbf{8x}    & \textbf{16x}             \\ \hline
Bibcubic & 0.01  & 0.01  & 0.01  & 0.01             \\ \hline
GF~\cite{he2010guided}       & 2.749 & 2.89  & 2.81  & 3.21             \\ \hline
TGV~\cite{7676324}      & 29.57 & 29.42 & 29.28 & 36.2             \\ \hline
RDN~\cite{zhang2018residual}      & 3.46  & 2.35  & 2.29  & \textbackslash{} \\ \hline
SRFBN~\cite{li2019feedback}    & 0.5   & 0.23  & 0.24  & \textbackslash{} \\ \hline
MSG~\cite{hui2016depth}      & 0.3   & 0.32  & 0.38  & 0.44             \\ \hline
DU-REAL~\cite{wang2020depth}  & 0.24  & 0.24  & 0.22  & 0.22             \\ \hline
DepthSR~\cite{guo2018hierarchical}  & 1.84  & 1.85  & 1.86  & 1.85             \\ \hline
Ours     & \textbf{0.18}  & \textbf{0.17}  & \textbf{0.15}  & \textbf{0.16}             \\ \hline
\end{tabular}
\label{tab4}
\end{table}

\section{Conclusions}
\label{conclusions}

In this paper, We propose a multi-scale progressive fusion network for depth map super-resolution. Our proposed method is a coarse to fine process, and mainly consists of three branches. The color encoder branch and the depth encoder branch are used to extract the hierarchic features from the HR color image and the LR depth map respectively. Then the reconstruction branch applies progressive fusion blocks to restored the HR depth map. Experiments show that our proposed method achieves the state-of-the-art performance on depth map SR with four different up-sampling scales. In the depth map SR task, our method benefits from the multi-scale progressive fusion mechanism to achieve good results. At the same time, our method still adapts to real world scenarios. And it shows that fine HR depth map can also be obtained for low-resolution depth maps with poor captured conditions.

\noindent \textbf{Discussion} Like other learning-based SR network, our method will generate biased results when there is a large area of depth errors and missing. Therefore, our method also requires some preprocessing under certain circumstances. In the future, we will pay attention to this issue and try to address this problem with a learning mechanism which is not affected by missing or error depth.


{\small
\bibliographystyle{ieee_fullname}
\bibliography{egbib}
}

\end{document}